\documentclass[letterpaper, 9 pt, conference]{ieeeconf} % Comment this line out if you need a4paper 

\IEEEoverridecommandlockouts % This command is only needed if you want to use the \thanks command
 
\usepackage{cite}
\usepackage{amsmath,amssymb,amsfonts}
\usepackage{algorithmic}
\usepackage{graphicx}
\usepackage{textcomp}
\usepackage{xcolor}
\def\BibTeX{{\rm B\kern-.05em{\sc i\kern-.025em b}\kern-.08em
    T\kern-.1667em\lower.7ex\hbox{E}\kern-.125emX}}
\usepackage{booktabs}
\usepackage{multirow}
\usepackage[super]{nth}
\usepackage{tikz}

\newcommand*\circledB[1]{\tikz[baseline=(char.base)]{
            \node[shape=circle,fill,inner sep=0.2pt] (char) {\textcolor{white}{#1}};}}
\usepackage{soul}

\usepackage{setspace}
\usepackage{cleveref}

\usepackage{fancyhdr}
\fancypagestyle{firstpage}{
  \fancyhf{}
  \chead{To appear at the 18th International Conference on Control, Automation, Robotics and Vision (ICARCV), December 2024, Dubai, UAE.}
  \cfoot{\thepage}
}

\title{\LARGE \bf
Embodied Neuromorphic Artificial Intelligence for Robotics: \\ Perspectives, Challenges, and Research Development Stack
\vspace{-0.1cm}}

\author{Rachmad Vidya Wicaksana Putra, Alberto Marchisio, Fakhreddine Zayer, Jorge Dias, Muhammad Shafique% <-this % stops a space
\thanks{Rachmad Vidya Wicaksana Putra and Alberto Marchisio are with eBrain Lab, Division of Engineering, New York University (NYU) Abu Dhabi, United Arab Emirates;
{e-mail: \{rachmad.putra, alberto.marchisio\}@nyu.edu}
}%
\thanks{Fakhreddine Zayer and Jorge Dias are with Khalifa University, United Arab Emirates;
{e-mail: \{fakhreddine.zayer, jorge.dias\}@ku.ac.ae}
}%
\thanks{Muhammad Shafique is the Director of eBrain Lab, Division of Engineering, New York University (NYU) Abu Dhabi, United Arab Emirates;
{e-mail: muhammad.shafique@nyu.edu}
}
}

\begin{document}

\maketitle
\pagestyle{plain}
\thispagestyle{firstpage}

%%%%%%%%%%%%%%%%%%%%%%%%%%%%%%%%%%%%%%%%%%%%%%%%%%%%%%%%%%%%%%%%%%%%%%%%%%%%%%%%
\begin{spacing}{0.95}
\begin{abstract}
Robotic technologies have been an indispensable part for improving human productivity since they have been helping humans in completing diverse, complex, and intensive tasks in a fast yet accurate and efficient way. 
Therefore, robotic technologies have been deployed in a wide range of applications, ranging from personal to industrial use-cases. 
However, current robotic technologies and their computing paradigm still lack embodied intelligence to efficiently interact with operational environments, respond with correct/expected actions, and adapt to changes in the environments. 
Toward this, recent advances in neuromorphic computing with Spiking Neural Networks (SNN) have demonstrated the potential to enable the embodied intelligence for robotics through bio-plausible computing paradigm that mimics how the biological brain works, known as ``\textit{neuromorphic artificial intelligence (AI)}''.  
However, the field of neuromorphic AI-based robotics is still at an early stage, therefore its development and deployment for solving real-world problems expose new challenges in different design aspects, such as accuracy, adaptability, efficiency, reliability, and security. 
To address these challenges, this paper will discuss how we can enable embodied neuromorphic AI for robotic systems through our perspectives: 
(P1) Embodied intelligence based on effective learning rule, training mechanism, and adaptability; 
(P2) Cross-layer optimizations for energy-efficient neuromorphic computing;
(P3) Representative and fair benchmarks;
(P4) Low-cost reliability and safety enhancements;
(P5) Security and privacy for neuromorphic computing; and
(P6) A synergistic development for energy-efficient and robust neuromorphic-based robotics.
Furthermore, this paper identifies research challenges and opportunities, as well as elaborates our vision for future research development toward embodied neuromorphic AI for robotics. 
\end{abstract}

%%%%%%%%%%%%%%%%%%%%%%%%%%%%%%%%%%%%%%%%%%%%%%%%%%%%%%%%%%%%%%%%%%%%%%%%%%%%%%%%
%%%%%%%%%%%%%%%%%%%%%%%%%%%%%%%%%%%%%%%%%%%%%%%%%%%%%%%%%%%%%%%%%%%%%%%%%%%%%%%%
\vspace{-0.1cm}
\section{Introduction}
\label{Sec_Intro}
\vspace{-0.1cm}

Robotic technologies have become critical and indispensable in everyday life for improving human productivity and quality of service, because robots have capabilities to complete diverse, complex, as well as intensive tasks faster, more accurate, and more efficient as compared to humans. 
Therefore, their deployments for solving real-world problems are widely spread from personal to industrial application use-cases. 
However, current robotic technologies and computing paradigm still lack ``\textit{intelligence}'', which is defined as the capability to interact with operational environments, correctly interpret the sensors' signals (stimulus), respond the stimulus with proper actions to accomplish the goals within the expected time frame, then learn the impact of the actions and continuously adapt to changes in uncontrolled/dynamic operational environments in an efficient manner~\cite{Ref_Bartolozzi_EmbodiedNeuroIntel_Nature22}.
Furthermore, apart from the above performance aspect (i.e., \textit{accuracy}, \textit{adaptability}, and \textit{efficiency}), these robots are also expected to perform robust processing against \textit{reliability} and \textit{security threats}, thus providing reliable outputs under diverse operational environments.
%.
These requirements can potentially be fulfilled by the neuromorphic computing paradigm through Spiking Neural Networks (SNNs)~\cite{Ref_Roy_SpikeMachineIntel_Nature19, Ref_Schuman_OpportunityNeuro_Nature22, Ref_Bartolozzi_EmbodiedNeuroIntel_Nature22} due to the following reasons.
\begin{itemize} 
    \item SNNs employ sparse spike-based data transmission and computation, thus enabling ultra-low power/energy consumption, including both the training and inference phases~\cite{Ref_Roy_SpikeMachineIntel_Nature19}\cite{Ref_Putra_FSpiNN_TCAD20}. 
    \item SNNs can employ diverse learning rules, such as the bio-plausible ones, hence enabling efficient training under both supervised and unsupervised learning settings~\cite{Ref_Putra_FSpiNN_TCAD20}. 
    These (unsupervised) learning capabilities are especially beneficial to make SNN-based systems adapt to dynamic operational environments~\cite{Ref_Putra_lpSpikeCon_IJCNN22}.  
    \item Recent advances in neuromorphic computing showed great success in achieving high accuracy, low latency, low memory footprint, and ultra-low power/energy consumption~\cite{Ref_Roy_SpikeMachineIntel_Nature19}\cite{Ref_Rathi_SNNsurvey_CSUR23}. 
\end{itemize}
All these benefits make neuromorphic computing with SNNs is a strong candidate as the computation model and technologies for enabling intelligence in resource-constrained robotic systems under diverse operational environments. 
In this paper, we refer an intelligence enabled by neuromorphic computing to as \textit{``neuromorphic artificial intelligence (AI)''} or simply \textit{``neuromorphic intelligence''}.

\vspace{0.1cm}
This paper delves into the intersection of neuromorphic computing and robotics, and specifically focusing on \textit{how to leverage neuromorphic AI-based approaches to achieve embodied intelligence in robotic systems}. 
First, this paper provides an overview of the field of neuromorphic AI-based robotics (\textbf{Section~\ref{Sec_NeuroRobots}}), highlighting its significance and potential impact. 
Then, it explores the principles and applications of neuromorphic computing with SNNs (\textbf{Section~\ref{Sec_SNNs}}), emphasizing their relevance in enabling embodied intelligence. 
Finally, drawing from our insights and experiences, we offer perspectives on the development of neuromorphic AI-based robotic systems (\textbf{Section~\ref{Sec_Perspectives}}). 
This includes identifying key challenges and opportunities, and outlining our vision for future research development aimed at advancing the realization of embodied neuromorphic AI for robotics.

\begin{figure}[t]
\centering
\includegraphics[width=0.95\linewidth]{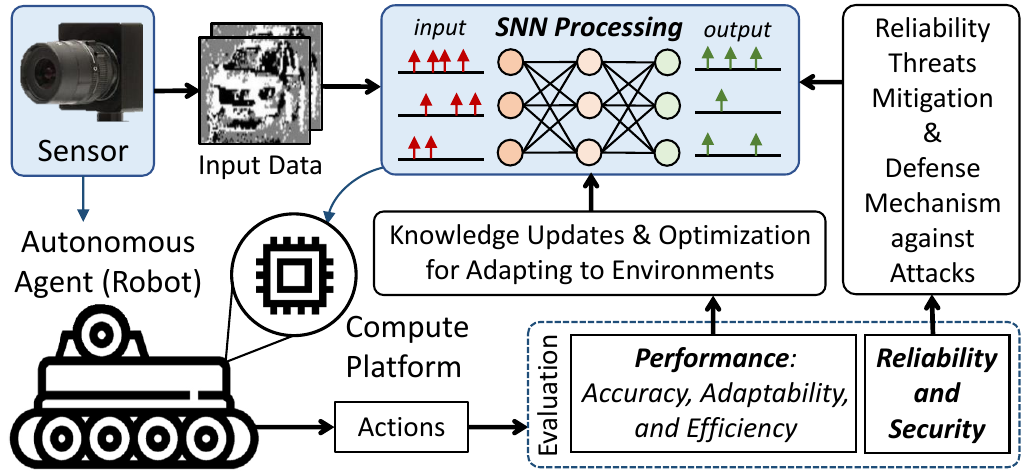}
\vspace{-0.3cm}
\caption{An overview of the embodied neuromorphic AI for robotics, which encompasses performance aspect like accuracy, adaptability, and efficiency, while considering reliability and security aspects.} 
\label{Fig_EmbodiedNeuroAI}
\vspace{-0.6cm}
\end{figure}

%%%%%%%%%%%%%%%%%%%%%%%%%%%%%%%%%%%%%%%%%%%%%%%%%%%%%%%%%%%%%%%%%%%%%%%%%%%%%%%%
%%%%%%%%%%%%%%%%%%%%%%%%%%%%%%%%%%%%%%%%%%%%%%%%%%%%%%%%%%%%%%%%%%%%%%%%%%%%%%%%
\section{Neuromorphic AI-based Robotics}
\label{Sec_NeuroRobots}

%%%%%%%%%%%%%%%%%%%%%%%%%%%%%%%%%%%%%%%
\subsection{System Overview}
\label{Sec_NeuroRobots_Overview}

\begin{figure*}[t]
\centering
\includegraphics[width=0.95\linewidth]{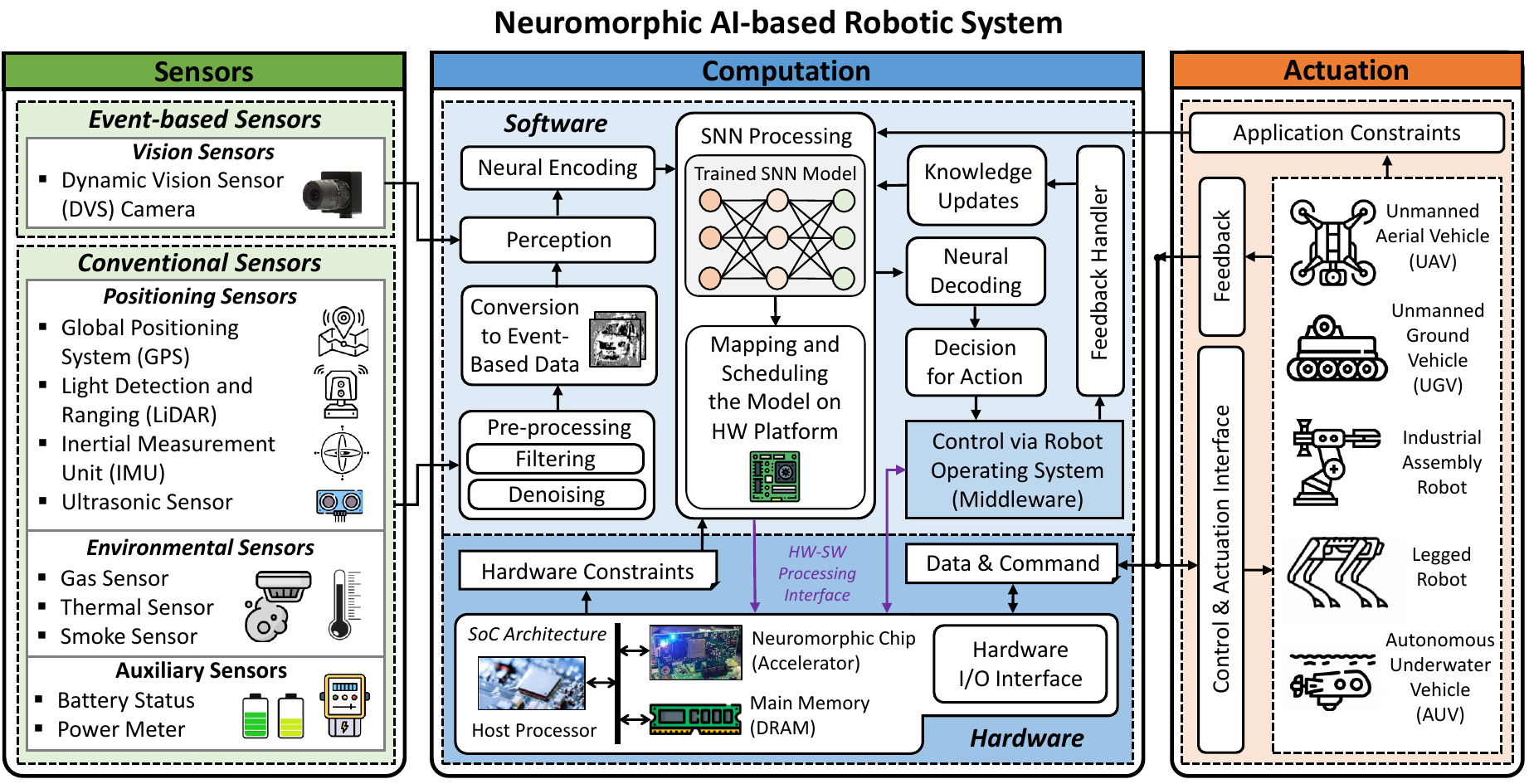}
\vspace{-0.2cm}
\caption{The system overview of the neuromorphic AI-based robotic system, showing the full robotics' processing pipeline.} 
\label{Fig_SystemOverview}
\vspace{-0.5cm}
\end{figure*}

Neuromorphic AI-based robotics integrates neuromorphic computing with SNNs to realize artificial intelligence~\cite{Ref_Rathi_SNNsurvey_CSUR23}. 
These robots need to autonomously interact with the environments~\cite{Ref_Bartolozzi_EmbodiedNeuroIntel_Nature22}, thereby necessitating efficient integration of the sensors, computation, and actuation modules, as shown in Fig.~\ref{Fig_SystemOverview}.

%%%%%%%%%%%% ---------
\vspace{0.1cm}
\subsubsection{\textbf{Sensors Subsystem}}
\label{Sec_NeuroRobots_Overview_Sensors}

This subsystem plays a pivotal role in the neuromorphic AI-based robotics for facilitating efficient interaction with the operational environments. 
These sensors encompass a wide range of input devices aimed at capturing diverse environmental data essential for the robots' autonomous operations.
Event-based sensors offer rapid data acquisition capabilities, thereby enabling the system to react swiftly to dynamic surroundings~\cite{Ref_Sironi_HATS_CVPR18}. 
Meanwhile, the conventional sensors such as Global Positioning System (GPS), Light Detection and Ranging (LiDAR), Inertial Measurement Unit (IMU) and ultrasonic, also contribute to precise robot positioning and navigation, which are crucial for navigating complex environments with high accuracy. 
Furthermore, environmental sensors such as thermal and gas detectors, provide valuable insights into the surroundings, ensuring the robots' adaptability to various conditions~\cite{Ref_Bartolozzi_EmbodiedNeuroIntel_Nature22}. 
Incorporating sophisticated data fusion techniques enhances the systems' ability to interpret and utilize complex environmental information effectively~\cite{Ref_Safa_CameraRadarSLAMsnn_ICRA23}. 
Furthermore, embracing multi-modality sensing and computing not only enriches the robots' perception capabilities but also enhances its overall robustness and adaptability~\cite{Ref_Bartolozzi_EmbodiedNeuroIntel_Nature22}\cite{Ref_Safa_CameraRadarSLAMsnn_ICRA23}. 

%%%%%%%%%%%% ---------
\vspace{0.1cm}
\subsubsection{\textbf{Computation Subsystem}}
\label{Sec_NeuroRobots_Overview_Compute}

This subsystem is responsible for performing the main SNN computation, which encompasses cross-layer design aspects: software (SW) part and hardware (HW) part.

\textbf{Software:}
If the input samples come in the form of events/spikes (e.g., from DVS camera), then spike/neural encoding is performed to provide interpretable sequences of spikes (i.e., \textit{spike trains}). 
Meanwhile, if the input samples come in the form of conventional analog or digital values, then data conversion to spikes is required. 
Such a conversion is typically coupled with neural encoding. 
For instance, a higher intensity image pixel is converted using a rate coding into a spike train with a higher spiking rate than the lower intensity one~\cite{Ref_Diehl_STDPmnist_FNCOM15}. 
Furthermore, we may need a pre-processing like denoising or filtering to ensure that the input samples are ideal for conversion and further processing~\cite{Ref_Marchisio_RSNN_IROS21}.   
The generated spike trains are then fed to the network model for main SNN processing. 
This processing is mapped and scheduled to run on the targeted hardware platform (e.g., \textit{neuromorphic HW accelerator}) while considering the application and hardware constraints. 
Outputs of this processing are interpreted based on the neural coding to derive an action or decision for the robot actuation.  
This processing is usually for inference, but can also be used for online training if the robotic system is designed to update its knowledge at run time for adapting to changes in the operational environments (i.e., \textit{continual learning})~\cite{Ref_Putra_SpikeDyn_DAC21}. 
Therefore, feedback from the actuation subsystem can be utilized for learning the impact of the previous action, and hence improving the adaptability.  

\textbf{Hardware:}
To efficiently expedite SNN processing, the robotic system needs to employ neuromorphic-based platform as the underlying hardware, since spike-based operations in SNNs are different from conventional Artificial Neural Network (ANN) operations~\cite{Ref_Basu_SNNicSurvey_CICC22}.
Hence, to meet this requirement, researchers in the neuromorphic community have proposed different neuromorphic accelerators/processors~\cite{Ref_Basu_SNNicSurvey_CICC22}. 
Furthermore, SNN processing should also be mapped and scheduled judiciously considering the dataflow of underlying hardware architecture to maximize the latency and energy efficiency benefits.      

%%%%%%%%%%%% ---------
\vspace{0.1cm}
\subsubsection{\textbf{Actuation Subsystem}}
\label{Sec_NeuroRobots_Overview_Control}

This subsystem has a pivotal role for realizing action (actuation) based on the given command.  
Therefore, application constraints should be communicated to the computation part, so that the generated response meets the requirements (e.g., maximum response time). 
Furthermore, to improve the adaptability of the robotic system, the impact of the previous action should be used as feedback for enriching and updating the robots' knowledge through a continual learning mechanism.  

%%%%%%%%%%%%%%%%%%%%%%%%%%%%%%%%%%%%%%%
\subsection{Current Trends of Neuromorphic-based Robotic Systems}
\label{Sec_NeuroRobots_Trends}

\begin{table*}
    \centering
    \caption{SNN Deployments in Robotic Applications}
    \label{Tab_StateOfTheArt}
    \vspace{-0.2cm}
    \scriptsize
    \begin{tabular}{|p{1.cm}|p{1.5cm}|p{2.5cm}|p{2.cm}|p{5.cm}| p{2.5cm}|}
        \hline
        \textbf{Work}& \textbf{Complexity}  & \multicolumn{1}{c|}{\textbf{Learning Algorithm}} & \multicolumn{1}{c|}{\textbf{Network}} & \multicolumn{1}{c|}{\textbf{Task-Specific}} & \multicolumn{1}{c|}{\textbf{HW Platform}} \\
        \hline
        \cite{Ref_Safa_CameraRadarSLAMsnn_ICRA23} &  SNN  & Continual STDP &  Feed-forward & Drone navigation & Laptop CPU \\
        \cite{Zhu10419072} & LSTM-SNN & Statistical learning & Recurrent & Conformal prediction for drone recognition & Laptop CPU \\
        \cite{10.3389/fnins.2021.667011} &  SNN & Wavefront & Feed-forward & Path planning & SpiNNaker \\
         \cite{Kreiser8594228} &  SNN & STDP & Feed-forward & SLAM & DYNAPSE, ROLLS\\
         \cite{Kreiser9197498} & CANN-SNN & STDP & Feed-forward & SLAM & Loihi\\ 
        \cite{zhang2023dynamic} & SNN & RL & Feed-forward & Object detection and obstacle avoidance & Laptop CPU \\      
        \cite{Yu10365576_2023} & SNN &  Threshold plasticity & Feed-forward & Adaptive tracking Control & Laptop CPU \\      
     \cite{Ref_Viale_CarSNN_IJCNN21} & SNN & STBP & Feed-forward &  Obstacle avoidance/Autonomous cars & Loihi \\
     \cite{BING202021} & SNN &  R-STDP, RL & Feed-forward &  Lane-keeping navigation  & Laptop CPU \\
     \cite{10.1007/978-3-319-46687-3_21} & SNN &  LSM & Reccurent &  End-effector tracking control & SpiNNaker \\
     \cite{doi:10.1126/sciadv.abl5068} & SNN &  Line-follower & No &  Navigation control & Mixed-A/D \\
        \hline
    \end{tabular}
    \label{Tab_StateOfTheArt}
    \vspace{-0.4cm}
\end{table*}

SNN deployment in robotics encompasses various applications, including path planning, simultaneous localization and mapping (SLAM), and adaptive control, among others. 
Path planning applications are addressed through diverse approaches such as feed-forward SNNs with Spike-Timing-Dependent Plasticity (STDP) for drone navigation~\cite{Ref_Safa_CameraRadarSLAMsnn_ICRA23}, LSTM-SNNs with statistical learning for drone recognition~\cite{Zhu10419072}, and SNNs with STDP for SLAM \cite{Kreiser8594228,Kreiser9197498}. 
Control applications utilize feed-forward SNNs with reinforcement learning (RL) for object detection and obstacle avoidance \cite{zhang2023dynamic}, SNNs with threshold plasticity for adaptive tracking control \cite{Yu10365576_2023}, and SNNs with reward-modulated STDP and RL for lane-keeping navigation \cite{BING202021}. 
This structured framework, summarized in Table~\ref{Tab_StateOfTheArt}, aids in selecting suitable SNN methods for specific robotic tasks, considering factors like network complexity and HW constraints, thus advancing robotics applications.

While significant progress has been made in the field of neuromorphic AI-based robotics, it remains at an early stage of development. 
Current state-of-the-art works primarily concentrate on enhancing learning quality (e.g., accuracy), but benchmarks are still limited. 
Moreover, little attention has been given to exploring reliability and security aspects to ensure robust processing in SNNs. Consequently, the absence of an end-to-end framework for developing energy-efficient and robust neuromorphic-based robots persists as a challenge.

%%%%%%%%%%%%%%%%%%%%%%%%%%%%%%%%%%%%%%%
\subsection{Open Challenges} 
\label{Sec_NeuroRobots_Challenges}

Based on the lessons learned from the current trends of the neuromorphic-based robotic systems, we identify multiple open challenges to be addressed for pushing the field forward and giving immediate and long-term impact in academics and industry. 
These challenges encompass a wide range of aspects, as the following. 
\begin{itemize}
    \item \textbf{Intelligence for Robotics:}
    This aspect demands robotic systems to have capabilities for accomplishing the goals (e.g., \textit{accuracy}) and adapting to changes in the operational environments (i.e., \textit{adaptability}).
    \item \textbf{Energy-Efficient Processing:}
    This aspect demands energy efficiency for SNN processing in the robotic system. 
    Therefore, it also demands low processing power and latency.
    \item \textbf{Limited Benchmarks for System Developments:} 
    Currently, we have limited benchmarks for supporting the developments of neuromorphic-based robotic systems, thereby limiting the quick progress of the field.
    \item \textbf{Reliability and Safety:}
    This aspect demands reliable processing in robotic systems in the presence of reliability and safety threats, thereby ensuring the correctness and fidelity of the generated output/decision.
    \item \textbf{Security and Privacy:}
    This aspect demands secure processing and privacy-preserving mechanism in robotic systems in the presence of security and privacy threats, thereby securing important and confidential information.
    \item \textbf{Limited Supports in Neuromorphic-based Robotic Developments:}
    Currently, there are limited supports (e.g., frameworks and tools) in developing robust and energy-efficient neuromorphic computing for robotics, hence limiting the adoption and deployment of neuromorphic-based robotic systems. 
\end{itemize}
These challenges and the corresponding opportunities are discussed further in Section~\ref{Sec_Perspectives} while elaborating our perspectives and visions.

%%%%%%%%%%%%%%%%%%%%%%%%%%%%%%%%%%%%%%%%%%%%%%%%%%%%%%%%%%%%%%%%%%%%%%%%%%%%%%%%
%%%%%%%%%%%%%%%%%%%%%%%%%%%%%%%%%%%%%%%%%%%%%%%%%%%%%%%%%%%%%%%%%%%%%%%%%%%%%%%%
\section{Neuromorphic Computing with SNNs}
\label{Sec_SNNs}

\begin{figure}[t]
\centering
\includegraphics[width=0.75\linewidth]{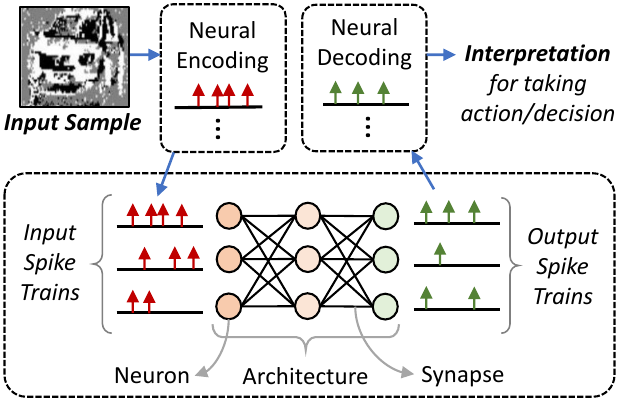}
\vspace{-0.3cm}
\caption{Overview of an SNN processing pipeline, showing different SNN components of such as the network architecture with neurons and synapses as well as neural coding (i.e., encoding and decoding).} 
\label{Fig_SNN}
\vspace{-0.5cm}
\end{figure}

SNNs offer high plausibility to the biological brain, since they employ spikes for transmitting information and spiking neurons for processing information. 
SNNs also have capabilities to learn spatial and temporal (spatio-temporal) information from the input samples.
In SNN processing, input samples are converted into input spike trains using a neural coding.
These input spikes are then fed to the SNN, which has a specific network architecture (topology). 
In each neuron, the input spikes will increase the neurons’ membrane potential, and if the potential reaches the defined threshold, then the respective neuron generates output spikes. 
Details of the neuronal dynamics may differ across different neuron models.
Afterward, the output spikes are decoded to interpret the information.
In the training phase, a specific learning rule is employed for extracting information from input features and updating the weights (i.e., knowledge) of the SNN model; while in the inference phase, the model infers input samples based on its learned knowledge.
In summary, an SNN model consists of several components: \textit{network architecture}, \textit{neuron model}, \textit{spike/neural coding}, and \textit{learning rule}~\cite{Ref_Putra_FSpiNN_TCAD20} (see an overview of the SNN processing pipeline in Fig.~\ref{Fig_SNN}), which are briefly discussed in the following.
\begin{itemize}
    \item \textbf{Network Architecture:}
    It defines how neurons are connected to each others via synapses. 
    There are several connection types, such as convolution, full-connected (dense), recurrent, or even combination from multiple architecture types~\cite{Ref_Rathi_SNNsurvey_CSUR23}.  
    \item \textbf{Neuron Model:}
    Several neuron models have been proposed in the literature, e.g., Hodgkin-Huxley, Integrate-and-Fire (IF), Leaky Integrate-and-Fire (LIF), and Izhikevich. 
    LIF neuron model is employed widely in the neuromorphic community as it can provide diverse spike representations like brain signals with a relatively low computational complexity~\cite{Ref_Putra_FSpiNN_TCAD20}; see LIF definition in Eq.~\ref{Eq_LIF}. 
    $V$ denotes the neurons' membrane potential, $V_r$ denotes the reset potential, $V_{th}$ denotes the threshold potential, $X$ denotes the input, and $\tau$ denotes time constant. 
    \begin{equation}
    \begin{split}
        \tau \frac{dV(t)}{dt} &  = - (V(t)-V_{r}) + X(t) \\
        \text{when} \;\;\; V & \geq V_{th} \;\;\; \text{then} \;\;\; V \leftarrow V_{r}
    \end{split}
    \label{Eq_LIF}
    \end{equation}
    \item \textbf{Spike/Neural Coding:} 
    It defines how information is represented in spikes (i.e., encoding), and how spikes are interpreted (i.e., decoding). 
    Several coding schemes have been proposed in the literature, such as rate (frequency), burst, rank order, and time-to-first spike~\cite{Ref_Rathi_SNNsurvey_CSUR23}. 
    \item \textbf{Learning Rule:} 
    It defines how the learning mechanism is performed during the training process. 
    Several learning rules have been proposed in the literature, and they employ different approaches which can be classified as the following.
    \begin{itemize}
        \item \textit{\underline{ANN-to-SNN Conversion}}:
        This approach performs training in non-spiking (ANN) domain using back-propagation techniques~\cite{Ref_Rueckauer_ANNtoSNN_FNINS17}. 
        Once the training phase is finished, the trained model is converted into an SNN model. 
        Therefore, this approach is typically performed in supervised learning settings.
        \item \textit{\underline{Direct SNN Training}}: 
        This approach performs training directly in spiking domain. 
        Currently, there are two prominent techniques: (1) approximation-based back-propagation known as surrogate gradient learning~\cite{Ref_Neftci_SurrogateSNNs_IEEEMSP19}; and (2) layer-wise local learning~\cite{Ref_Kaiser_DECOLLE_FNINS20}. 
        This approach is performed under supervised learning settings to reduce the loss function, hence achieving state-of-the-art accuracy for SNNs.
        \item \textit{\underline{Bio-Plausible Learning}}:
        This approach also performs training directly in spiking domain using techniques inspired by biological phenomena in the brains, like Spike-Timing-Dependent Plasticity (STDP) learning rule~\cite{Ref_Putra_FSpiNN_TCAD20}.
        These rules perform learning locally in each synapse without having to know the labels, hence supporting unsupervised learning settings. 
        In this manner, these rules enable efficient \textit{unsupervised continual learning} where the robotics gather (unlabeled) data directly from the environments, learn from them, and update the knowledge accordingly~\cite{Ref_Putra_lpSpikeCon_IJCNN22}~\cite{Ref_Putra_SpikeDyn_DAC21}.
    \end{itemize}
\end{itemize}

%%%%%%%%%%%%%%%%%%%%%%%%%%%%%%%%%%%%%%%%%%%%%%%%%%%%%%%%%%%%%%%%%%%%%%%%%%%%%%%%
%%%%%%%%%%%%%%%%%%%%%%%%%%%%%%%%%%%%%%%%%%%%%%%%%%%%%%%%%%%%%%%%%%%%%%%%%%%%%%%%
\section{Our Perspectives}
\label{Sec_Perspectives}
\vspace{-0.1cm}

We now present our perspectives regarding new challenges and opportunities introduced by the neuromorphic AI-based robotics in several aspects which are correlated closely, as the following. 
\begin{itemize}
    \item[\textbf{P1:}] \textbf{Embodied Neuromorphic Intelligence for Robotics:} 
    \textit{``Enabling neuromorphic intelligence in robotic systems requires novel design methods beyond the traditional ones which rely on the offline training and the SNN model deployment without any knowledge updates''}.
    \item[\textbf{P2:}] \textbf{HW/SW-level Optimizations for Energy-Efficient Neuromorphic Computing:} 
    \textit{``Maximizing energy efficiency of the neuromorphic computing in robotics needs to consider cross-layer optimizations covering both the HW and SW layers that complement each other''}.
    \item[\textbf{P3:}] \textbf{Representative and Fair Benchmarks for Robotics:} 
    \textit{``It is required to provide representative and fair benchmarks for robotic developments that cover different sensory functions and objective tasks under diverse environment settings''}.
    \item[\textbf{P4:}] \textbf{Low-Cost Reliability and Safety Enhancements:}
    \textit{``Low-cost techniques are required to enhance the reliability and safety of the neuromorphic-based robotic systems, i.e., reliability against device faults to ensure safety of robots' behavior''}.
    \item[\textbf{P5:}] \textbf{Security and Privacy for Neuromorphic Computing:}
    \textit{``Cost-effective techniques are needed to ensure the security and privacy of the neuromorphic-based robotic systems, i.e., security against attacks and privacy-preserving mechanism''}.
    \item[\textbf{P6:}] \textbf{A Synergistic Development for Energy-Efficient and Robust Neuromorphic-based Robotic Systems:}
    \textit{``It is important to have an end-to-end development framework and tools for the full processing pipeline of a robotic system (i.e., sensors, computation, and actuation) across the HW and SW layers''}.
\end{itemize}
In the following sub-sections, we elaborate these perspectives in more depth while providing some early evidences supporting our visions on how to move forward with the opportunities.

%%%%%%%%%%%%%%%%%%%%%%%%%%%%%%%%%%%%%%%
\subsection{\textbf{P1: Embodied Neuromorphic Intelligence for Robotics}} 
\label{Sec_Perspective_Intelligence}

Realizing the neuromorphic intelligence in robots needs the capabilities to accomplish the \textit{targeted learning quality} (e.g., accuracy or precision) and the \textit{adaptability} to interact with the environments. 

\vspace{0.1cm}
\textbf{Learning Quality:} 
To accomplish the targeted learning quality, a robot requires to perform appropriate actions by relying on its perception and existing knowledge. 
The targeted quality typically depends on the application use-case, because different use-cases often require (1) different \textit{data types} from sensors for perception, (2) different \textit{objective tasks} to solve through appropriate training mechanism and learning rule, and (3) different \textit{expected actions}. 
Therefore, in our perspective, it is important to design a specialized SNN processing pipeline for each application use-case to achieve the targeted learning quality through a synergistic design and/or selection of network architecture, learning rule, and neural coding. 
Center of this idea is to develop high quality learning rule and training mechanism that maximally learn the spatio-temporal information from the input samples. 
Here, multi-modality sensing will enrich the input information for guiding the training process better.

\begin{figure}[hbtp]
\vspace{-0.2cm}
\centering
\includegraphics[width=\linewidth]{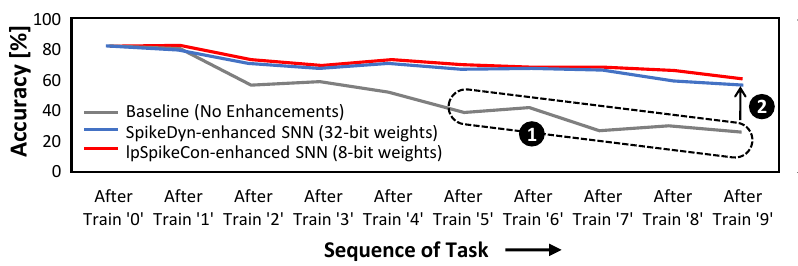}
\vspace{-0.7cm}
\caption{SNN accuracy in the unsupervised continual learning settings on the MNIST dataset, across different precision levels and methods:  Baseline~\cite{Ref_Diehl_STDPmnist_FNCOM15}, SpikeDyn~\cite{Ref_Putra_SpikeDyn_DAC21}, and lpSpikeCon~\cite{Ref_Putra_lpSpikeCon_IJCNN22}; adapted from studies in~\cite{Ref_Putra_lpSpikeCon_IJCNN22}. Here, samples from each task (class) are fed to the SNN and learned sequentially to simulate dynamic environments, then the SNN is expected to learn new tasks without facing \textit{catastrophic forgetting}, i.e., forgetting the previously learned knowledge (tasks) after learning new information.} 
\label{Fig_ContinualLearn}
\vspace{-0.2cm}
\end{figure}

\vspace{0.1cm}
\textbf{Adaptability:}
The application use-case also defines the expected level of adaptability of the robot. 
For instance, a robot that operates in a controlled environment and execute repetitive actions (e.g., an industrial assembly robot) requires a lower adaptability level as compared to the one that has to continuously infer and adapt to dynamically-changing operational environments, including changes in the robotic platform itself (e.g., battery and thermal status). 
Such adaptive capabilities are important due to the following reasons. 
\begin{itemize}
    \item Some application use-cases are physically too far away and/or dangerous for humans to explore (e.g., space, active volcano, deep forest, and open sea), thereby requiring smart autonomous agents/robots that can adapt to the unseen conditions~\cite{Ref_Putra_Mantis_ICARA23}.
    \item The knowledge learned from the offline training may become obsolete over time, which leads to low accuracy at run time in dynamic environments; see \circledB{1} in Fig.~\ref{Fig_ContinualLearn}(a).
    \item Dynamic environments often have unseen data that need to be learned online. 
    Hence, the robot requires knowledge updates through \textit{online training (learning)} by leveraging unlabeled data that are gathered directly from the environments~\cite{Ref_Putra_lpSpikeCon_IJCNN22}.
    \item Robots also need to trade-off their performance (e.g., accuracy and latency) with processing power/energy for saving battery. 
\end{itemize}

Recent progress has demonstrated the SNN potentials for overcoming \textit{unsupervised continual learning} challenges. 
For instance, recent works~\cite{Ref_Putra_SpikeDyn_DAC21}\cite{Ref_Putra_lpSpikeCon_IJCNN22} improve the accuracy of baseline network (with 32-bit weights and STDP rule) using learning rule enhancements; see \circledB{2} in Fig.~\ref{Fig_ContinualLearn}(b).
The work of \cite{Ref_Putra_SpikeDyn_DAC21} enables unsupervised continual learning by leveraging spiking activities to enrich the STDP learning process~\cite{Ref_Putra_SpikeDyn_DAC21}. 
Then, this concept is further extended in \cite{Ref_Putra_lpSpikeCon_IJCNN22} by employing a guided fine-tuning on different learning parameters to enable network with lower weight precision to achieve comparable accuracy to the ones with higher weight precision~\cite{Ref_Putra_lpSpikeCon_IJCNN22}.   
However, the state-of-the-art works still consider the MNIST dataset for evaluating their accuracy~\cite{Ref_Putra_lpSpikeCon_IJCNN22}. 
Therefore, in our perspective, it is important to develop high quality learning rules and training mechanism that can overcome diverse environments, by leveraging unlabeled data gathered during run time for updating the knowledge. 

\textit{\underline{Research developments and opportunities:}} 
To summarize, we list new opportunities that arise from from the requirements of embodied neuromorphic intelligence for robotics.
\begin{itemize}
    \item A specialized SNN processing pipeline should be developed for each application use-case to achieve high learning quality with advanced SNN architectures and learning techniques.
    \item Proper learning rule and training mechanism are the keys to effectively translate perception into a series of correct actions.  
    \item Embracing multi-modality sensing is important to enrich the knowledge for achieving better learning quality, and for adjusting the trade-off benefits between quality (e.g., high accuracy or precision) with efficiency (e.g., long battery lifespan).
    \item Efficient online training mechanism coupled with an unsupervised continual learning is a potential approach for making the robots adaptive and capable of solving unseen conditions without having to be explicitly programmed.
    \item Neuromorphic intelligence is highly potential for fostering the emerging robotic applications, such as the following. 
    \begin{itemize}
        \item Swarm intelligence with advanced algorithms for coordinating multiple neuromorphic-based robots to work together efficiently in tasks like search and rescue or monitoring.
        \item Miniaturization of neuromorphic-based robots with embedded flexible electronics for distributed computation, enabling them to operate in tight spaces and perform agile tasks.
        \item Fusion of neuromorphic-based robotics with Brain-Machine Interfaces (BMIs) to create advanced interfaces for controlling robots using brain activity.
    \end{itemize}
\end{itemize}

%%%%%%%%%%%%%%%%%%%%%%%%%%%%%%%%%%%%%%%
\subsection{\textbf{P2: HW/SW-level Optimizations for Energy-Efficient Neuromorphic Computing}} 
\label{Sec_Perspective_Platforms}

Energy efficiency is crucial for battery-powered robotic systems to enable a long service time. 
This necessitates HW and SW layers to be working together in an efficient manner.
To achieve this, optimizations in the HW and SW layers should be performed in a way that they complement each other; otherwise, the optimization benefits will be sub-optimal. 
Therefore, in our perspective, it is important to perform conjoint HW/SW-level optimizations for maximizing energy efficiency of neuromorphic-based robotics. 

\begin{figure}[hbtp]
\centering
\includegraphics[width=0.925\linewidth]{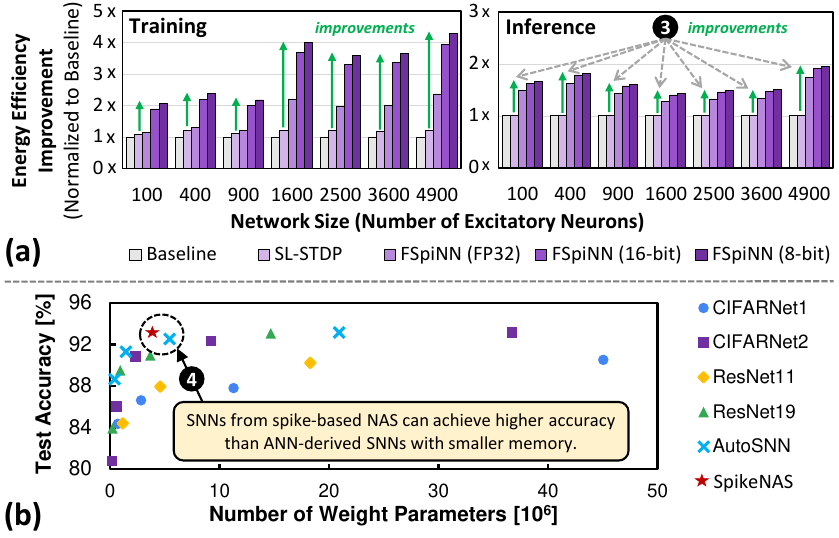}
\vspace{-0.3cm}
\caption{\textbf{(a)} Neuron elimination and quantization in the FSpiNN~\cite{Ref_Putra_FSpiNN_TCAD20} improve energy efficiency from the SL-STDP~\cite{Ref_Srinivasan_SLSTDP_IJCNN17} and the baseline model~\cite{Ref_Diehl_STDPmnist_FNCOM15}; adapted from studies in~\cite{Ref_Putra_FSpiNN_TCAD20}. 
\textbf{(b)} Accuracy and memory footprints of different SNNs for the CIFAR10 dataset: ResNet11, ResNet19, CIFARNet1, CIFARNet2, AutoSNN, and SpikeNAS; adapted from studies in~\cite{Ref_Putra_SpikeNAS_arXiv24}.
} 
\label{Fig_FSpiNN_NAS}
\vspace{-0.2cm}
\end{figure}

\vspace{0.1cm}
\textbf{SW-level Optimizations:} 
To decrease the compute and memory requirements of SNNs, model compression approach is usually employed, as it effectively reduces the memory footprint by introducing sparsity in weights and/or neurons. 
This approach includes techniques such as weight pruning, neuron elimination and quantization~\cite{Ref_Putra_FSpiNN_TCAD20}\cite{Ref_Putra_QSpiNN_IJCNN21}. 
Another approach is optimizing the computational time of neuron operations through approximation~\cite{Ref_Sen_ApproxSNN_DATE17} and timestep reduction~\cite{Ref_Chowdhury_LowLatencySNNs_ECCV22}\cite{Ref_Putra_TopSpark_IROS23}. 
Energy efficiency improvements through neuron elimination and quantization are shown by \circledB{3} in Fig.~\ref{Fig_FSpiNN_NAS}(a).
Furthermore, another potential approach is to build a network through neural architecture search (NAS)~\cite{Ref_Na_AutoSNN_ICML22, Ref_Kim_SNASNet_ECCV22, Ref_Putra_SpikeNAS_arXiv24, Ref_Achararit_APNAS_Access20}; see \circledB{4} in Fig.~\ref{Fig_FSpiNN_NAS}(b).
Moreover, incorporating HW constraints (e.g., memory budget) into the NAS process can guide the search to find a better network architecture~\cite{Ref_Putra_SpikeNAS_arXiv24}. 
Apart from model optimization, efficiency improvements can also be obtained by maximizing data reuse in the SNN processing dataflow considering the underlying HW platform. 
It aims at minimizing data movements, especially between off-chip memory and on-chip processor, as memory accesses usually dominate the systems' energy consumption~\cite{Ref_Krithivasan_SpikeBundle_ISLPED19}\cite{Ref_Putra_SparkXD_DAC21}. 

\begin{figure}[t]
\centering
\includegraphics[width=0.925\linewidth]{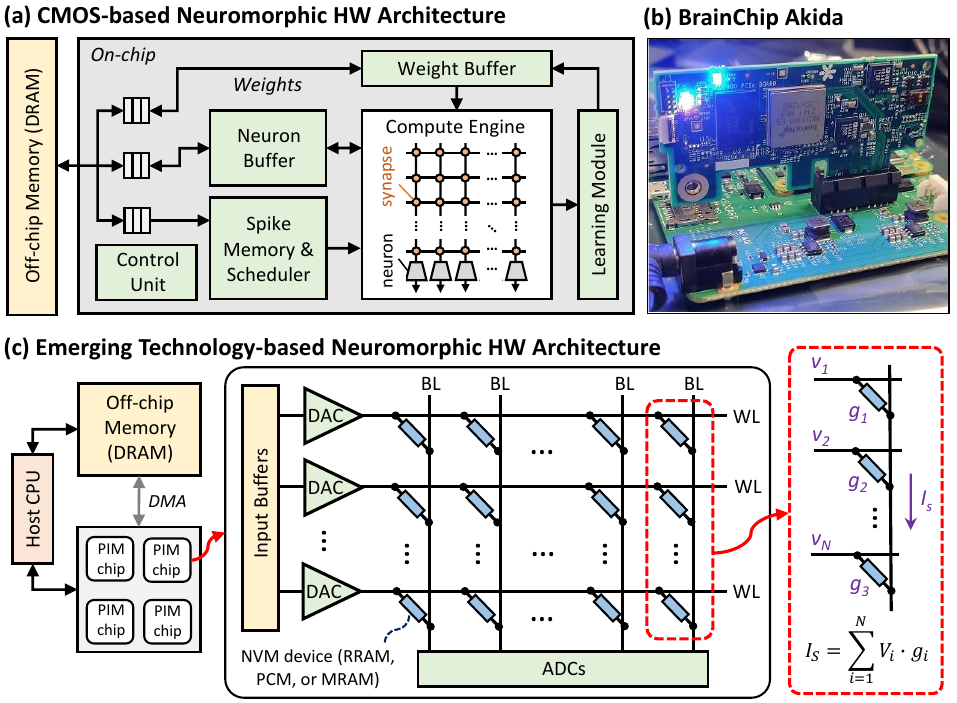}
\vspace{-0.3cm}
\caption{\textbf{(a)} The typical CMOS-based single-core neuromorphic architecture; adapted from studies in~\cite{Ref_Putra_SoftSNN_DAC22}. \textbf{(b)} The hardware setup of the BrainChip Akida neuromorphic processor in our lab. \textbf{(c)} The typical emerging technology-based neuromorphic architecture; adapted from studies in~\cite{Ref_Asifuzzaman_SurveyPIM_Memori23}.} 
\label{Fig_HWplatforms}
\vspace{-0.5cm}
\end{figure}

\textbf{HW-level Optimizations:} 
To expedite SNN processing as well as maximize its energy efficiency, researchers have proposed neuromorphic HW accelerators that enable efficient spike transmission and computation~\cite{Ref_Basu_SNNicSurvey_CICC22}; see Fig.~\ref{Fig_HWplatforms}(a). 
Then, employments of optimized memories (e.g., approximate memories~\cite{Ref_Putra_EnforceSNN_FNINS22}) can add further efficiency benefits, because memory access is significantly more expensive than arithmetic operations~\cite{Ref_Putra_DRMap_DAC20}\cite{Ref_Putra_ROMANet_TVLSI21}.
Besides CMOS-based technologies, researcher also explored non-volatile memory (NVM) technologies (e.g., RRAM: Resistive Random Access Memory, MRAM: Magnetic RAM, and PCM: Phase Change Memory) for realizing processing-in-memory (PIM) or compute-in-memory (CIM) paradigm which minimizes data movements, thus reducing energy consumption~\cite{Ref_Asifuzzaman_SurveyPIM_Memori23}. 
The typical PIM architecture is shown in Fig.~\ref{Fig_HWplatforms}(b).
Furthermore, the run-time power managements can also be employed to improve energy efficiency, such as clock gating, power gating, and dynamic voltage frequency scaling (DVFS)~\cite{Ref_Shafique_EdgeAI_ICCAD21}.

\vspace{0.1cm}
\textit{\underline{Research developments and opportunities:}} 
To summarize, we list new opportunities that arise from the requirements of HW/SW-level optimization techniques.
\begin{itemize}
    \item Optimization techniques that have been developed in the artificial neural network (ANN) domain can be leveraged for SNNs, but they have to be crafted specifically for SNN operations. 
    \item Conjoint cross-layer optimizations for both HW and SW layers have to complement each other to maximize the optimization benefits from both layers. 
    \item Developments of specialized neuromorphic accelerators (with CMOS and/or emerging technologies) are crucial to maximally improve energy efficiency of SNN processing for robotics. 
    Here, run-time power management modes can also be incorporated for further power/energy optimization. 
    \item In some use-cases, neuromorphic HW architectures should be designed to facilitate both the training and inference phases, to enable efficient online training (learning) for updating the knowledge of robotic systems at run time. 
\end{itemize}

%%%%%%%%%%%%%%%%%%%%%%%%%%%%%%%%%%%%%%%
\subsection{\textbf{P3: Representative and Fair Benchmarks for Robotics}}
\label{Sec_Perspective_Benchmarks}

\begin{figure}[h]
\vspace{-0.2cm}
\centering
\includegraphics[width=\linewidth]{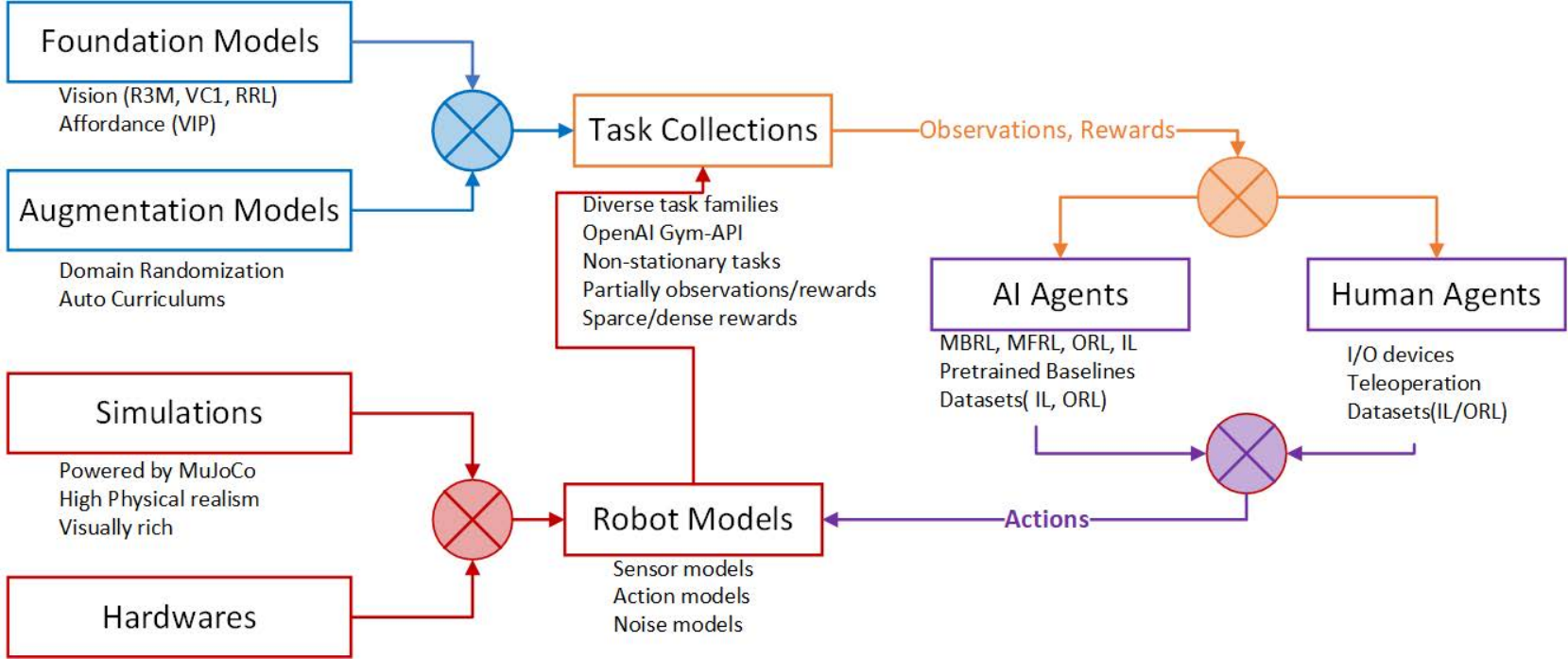}
\vspace{-0.6cm}
\caption{RoboHive's adaptability stems from its modular design, featuring task collections, diverse agents, a unified robot for simulation and reality, and an augmentation module; adapted from studies in~\cite{kumar2024robohive}.} 
\label{Fig_Robohive}
\vspace{-0.2cm}
\end{figure}

Representative and fair benchmarks for robotics are developed across various frameworks and initiatives. 
Within ROS, benchmarks encompass tasks like object recognition, navigation, manipulation, and cover diverse environments. 
OpenAI Gym offers benchmark environments for reinforcement learning tasks, including robotic arm manipulation and locomotion in complex settings~\cite{brockman2016openai}. 
Robotics challenges such as the DARPA Robotics Challenge and RoboCup provide scenarios for testing perception, manipulation, and interaction abilities in dynamic environments~\cite{krotkov2018darpa}. 
Meanwhile, research labs and consortia focus on specific domains such as object manipulation and human-robot interaction, offering tailored benchmarks for these areas~\cite{Andrew2013low}. 
Additionally, platforms like RoboHive facilitate collaborative robotic research and development, for contributing to the creation of standardized benchmarks and evaluation metrics within the community~\cite{kumar2024robohive}, as outlined in Fig.~\ref{Fig_Robohive}.

Rich sensory information is crucial for various functions. For vision, datasets like DAVIS and ATIS offer event-based visual data, enabling tasks such as object recognition and motion tracking~\cite{Gallego_2022TPAMI_EventBasedVisionSurvey}. 
Auditory perception benefits from datasets like DVS-DAVIS, incorporating event-based audio-visual data for sound localization~\cite{schoepe2023event}. 
Touch sensing datasets like Prophesee's TacTip dataset provide event-based tactile data, aiding in understanding object properties~\cite{muthusamy2020neuromorphic}. 
Position sensing datasets from GPS, LiDAR, IMU, and ultrasonic sensors offer event-based information for localization and mapping tasks~\cite{cadena2016past}. 
Environmental sensing datasets capture gas, thermal, and smoke events, enabling robots to detect environmental changes~\cite{xing2019firenose}. Auxiliary sensing datasets, including battery status and power meter data, help manage energy resources efficiently~\cite{hasib2021comprehensive}. 
Together, these datasets facilitate the development of neuromorphic robotic systems capable of perceiving and interacting with the environment effectively.

\vspace{0.1cm}
\textit{\underline{Research developments and opportunities:}} 
To summarize, we list new opportunities from the benchmark suite aspect.
\begin{itemize}
    \item Each robotic application use-case should have a well-defined benchmarks, including functionality (e.g., accuracy or precision), memory footprint, latency (response time), power/energy consumption, and figure of merit like synaptic operations-per-second (SOPS).
    \item Integrated benchmarking tools are required for comprehensively evaluating the performance of SNN processing in the neuromorphic-based robots for executing the given tasks. 
    \item Ethical implications of the neuromorphic-based robots should be investigated, and guidelines for their responsible development and deployment should be established as well.
\end{itemize}

%%%%%%%%%%%%%%%%%%%%%%%%%%%%%%%%%%%%%%%
\subsection{\textbf{P4: Low-Cost Reliability and Safety Enhancements}} 
\label{Sec_Perspective_Reliability}

Modern nano-scale computing platforms like neuromorphic processors in robotic systems are vulnerable to reliability issues. 
Common reliability threats include \textit{permanent faults} due to process variations during the chip manufacturing process~\cite{Ref_Putra_ReSpawn_ICCAD21}\cite{Ref_Putra_RescueSNN_FNINS23}, \textit{transient faults (soft errors)} in form of bit flips due to high-energy particle strikes~\cite{Ref_Putra_SoftSNN_DAC22}, and \textit{aging} due to device/circuit wear out over time~\cite{Ref_Shafique_EdgeAI_ICCAD21}. 
These HW faults can appear in both memories and compute units, inducing data errors that may propagate to application layer and affect the actions of robotic systems, such as threatening the \textit{safety} of robotics' behavior. 

\begin{figure}[hbtp]
\vspace{-0.2cm}
\centering
\includegraphics[width=\linewidth]{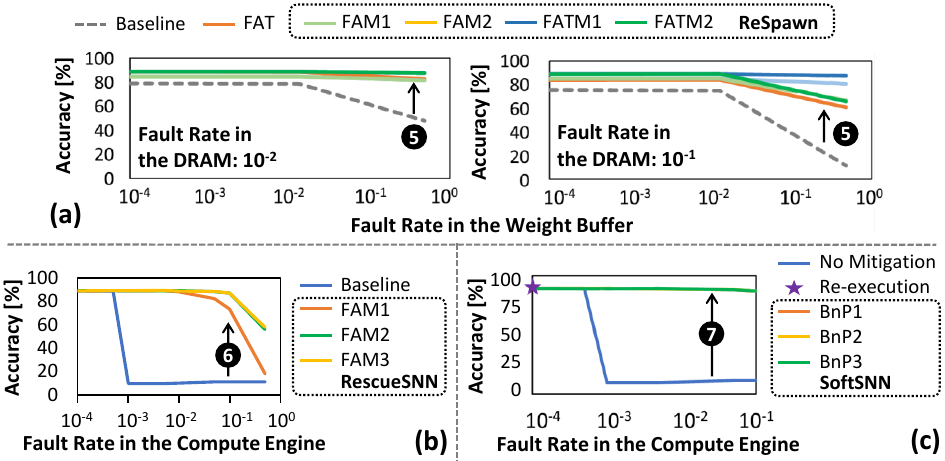}
\vspace{-0.6cm}
\caption{SNN accuracy of different techniques for mitigating HW faults in the neuromorphic accelerator on the MNIST. 
\textbf{(a)} Permanent fault mitigation in the weight memories; adapted from studies in~\cite{Ref_Putra_ReSpawn_ICCAD21}. 
\textbf{(b)} Permanent fault mitigation in the compute engine; adapted from studies in~\cite{Ref_Putra_RescueSNN_FNINS23}. 
\textbf{(c)} Soft error mitigation in the compute engine; adapted from studies in~\cite{Ref_Putra_SoftSNN_DAC22}.} 
\label{Fig_FaultMitigation}
\vspace{-0.3cm}
\end{figure}

Therefore, in our perspective, it is important to mitigate such reliability threats in a cost-effective manner thereby preventing the systems' dysfunctionality even in the presence of errors~\cite{Ref_Putra_ReSpawn_ICCAD21, Ref_Putra_RescueSNN_FNINS23, Ref_Putra_SoftSNN_DAC22}.
Recent progress has demonstrated that these reliability threats can be mitigated, as shown in Fig.~\ref{Fig_FaultMitigation}. 
Specifically, the recent work mitigates permanent faults in the weight memories by employing Fault-Aware Training (FAT), Fault-Aware Mapping (FAM), and Fault-Aware Training-and-Mapping (FATM) techniques~\cite{Ref_Putra_ReSpawn_ICCAD21}; see \circledB{5}. 
To mitigate permanent faults in the compute engine, the recent work employs Fault-Aware Mapping (FAM) techniques~\cite{Ref_Putra_RescueSNN_FNINS23}; see \circledB{6}.
Meanwhile, the recent work mitigates soft errors in the compute engine, by employing Re-execution and Bound-and-Protect (BnP) techniques~\cite{Ref_Putra_SoftSNN_DAC22}; see \circledB{7}.
However, these works have not considered robotic datasets in their training phase.
Hence, we can leverage these fault mitigation techniques for robotics by considering the workloads and specific characteristics of the robotic applications. 

\vspace{0.1cm}
\textit{\underline{Research developments and opportunities:}} 
To summarize, we list new opportunities from reliability and safety aspects.
\begin{itemize}
    \item Design and implementation of advanced reliability and safety threats are essential to explore possible reliability and safety issues, as well as their respective countermeasures.
    \item Cost-effective fault mitigation techniques should be applied in the robots, to ensure reliable and safe SNN executions in the presence of faulty devices and enable long battery lifespan.
    \item Reliability and safety should be considered as key optimization objectives in a joint multi-objective design of accurate, efficient, as well as reliable and safe neuromorphic-based robots.
\end{itemize}

%%%%%%%%%%%%%%%%%%%%%%%%%%%%%%%%%%%%%%%
\vspace{-0.1cm}
\subsection{\textbf{P5: Security and Privacy for Neuromorphic Computing}} 
\label{Sec_Perspective_Security}

\begin{figure}[hbtp]
\vspace{-0.3cm}
\centering
\includegraphics[width=0.92\linewidth]{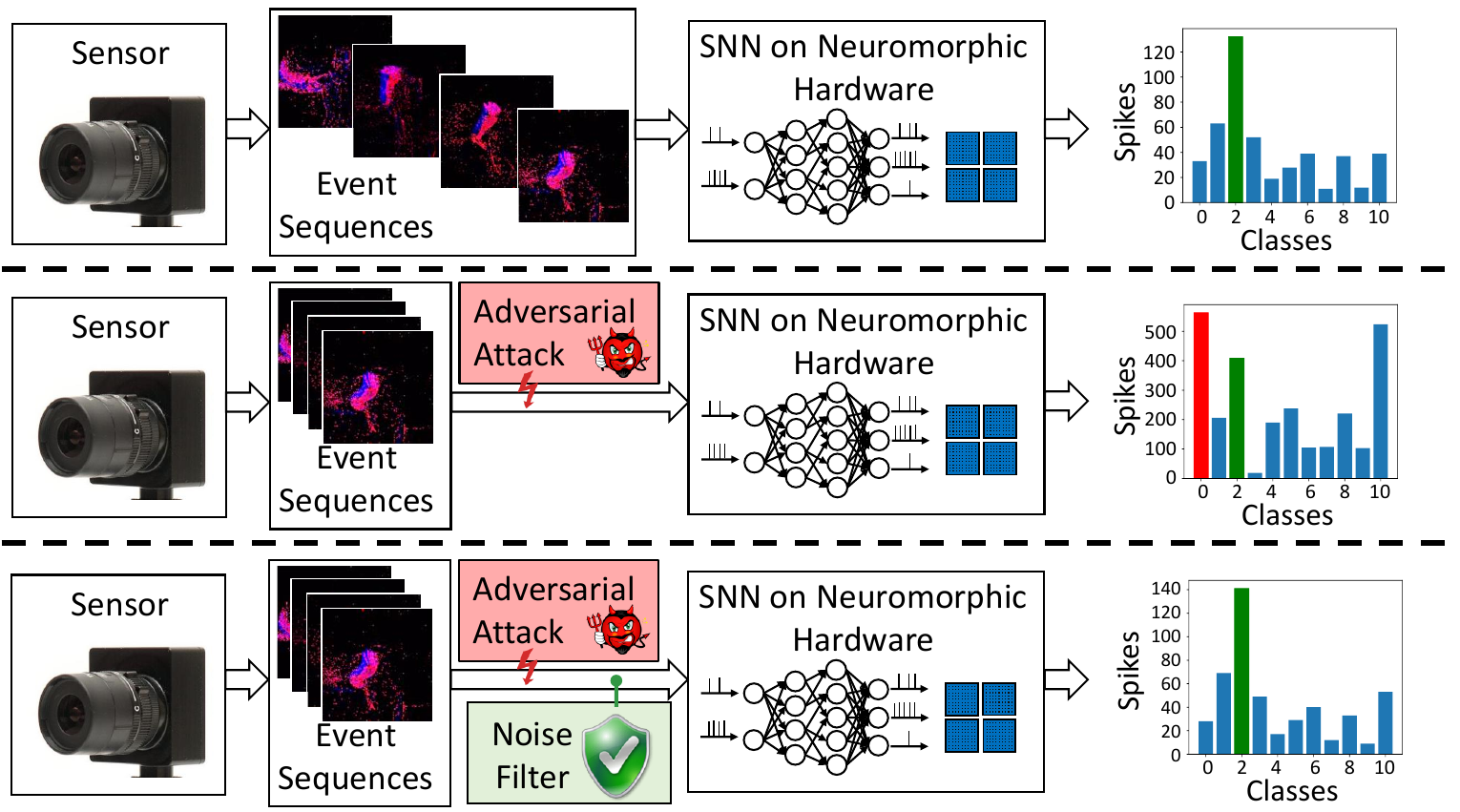}
\vspace{-0.3cm}
\caption{Example showing event-based adversarial attacks applied to gesture recognition systems implemented on neuromorphic hardware, in which a noise filter is applied to improve the SNN robustness against adversarial attacks; adapted from studies in~\cite{Marchisio_2021IROS_RSNN}.} 
\label{Fig_DVS_Attacks_Filter}
\vspace{-0.2cm}
\end{figure}

Ensuring the correct functionality of SNNs in the presence of adverse conditions is fundamental for safety-critical applications. 
Such adverse conditions include both \textit{naturally noisy environments} (e.g., fog, rain, and snow) and \textit{malicious agents} that deliberately aim to fool the systems or infer confidential data. 
Malicious agent conditions can either be identified as \textit{backdoor attacks} which aim to fool the system during the training phase), \textit{adversarial attacks} which alter the systems' correct functionality during the inference phase, or \textit{privacy attacks} which extract sensitive features of the training data or private information of the SNN model.
Studies in~\cite{Ref_ElAllami_SecuringSNNs_DATE21} suggest that SNNs offer higher intrinsic robustness against large adversarial attack perturbations than their counterpart Convolutional Neural Networks (CNNs). However, further security and privacy enhancements need to be designed.
To counter these adversities, several defense methodologies have been proposed~\cite{Ref_ElAllami_SecuringSNNs_DATE21, Marchisio_2021IROS_RSNN, Kim_2022AAAI_PrivateSNN, Nikfam_2023Information_HESNN}. An example of successful defense applied to event-based adversarial attacks is shown in Fig.~\ref{Fig_DVS_Attacks_Filter}. From the results, it is evident that the adversarial attack fools the system, but the noise filter removes the effect of the perturbation and allows a correct SNN classification.

From our perspective, it is important to enhance the robustness not only against environmental noises, but also against a large variety of vulnerability threats~\cite{Ref_Marchisio_IsSpikingSecure_IJCNN20, Ref_Venceslai_NeuroAttack_IJCNN20, Marchisio_2021IJCNN_DVSAttacks}, to minimize the chances that malicious adversaries unveil unseen attack scenarios. 
Moreover, due to the stringent resource and memory constraints required in mobile robotic systems, it is essential that the security countermeasures are lightweight and do not overload the systems with expensive computations and long execution times, as they would severely limit their applicability in practical use-cases~\cite{Ref_Shafique_EdgeAI_ICCAD21}\cite{Dave_2022VTS_AgileMethodology}.
Furthermore, it is also crucial to design and optimize networks for security and privacy from the start of the design phase~\cite{Marchisio_2022Access_RoHNAS}, since post-design robustness optimizations might not be as effective. 

\textit{\underline{Research developments and opportunities:}} 
To summarize, we list new opportunities from security and privacy aspects.
\begin{itemize}
    \item Design and implementation of advanced security attacks are essential to explore possible security threats and their respective countermeasures.
    \item Cost-effective security defenses should be designed to protect the robotic systems in the presence of adversaries, without incurring significant computation and resource overheads.
    \item Security and privacy should be considered as key optimization objectives in a joint multi-objective design of accurate, efficient, and secure neuromorphic-based robots.
\end{itemize}

%%%%%%%%%%%%%%%%%%%%%%%%%%%%%%%%%%%%%%%
\subsection{\textbf{P6: A Synergistic Development for Enabling Energy-Efficient and Robust Neuromorphic-based Robotic Systems}} 
\label{Sec_Perspective_Synergy}

The neuromorphic AI-based robotics field is still at an early stage, hence its developments targeting different design objectives (e.g., performance, reliability, and security) are still developed separately. 
Therefore, making it challenging to build a complete neuromorphic-based robotic system that offers high efficiency and robustness. 
The reason is that, developing a system with a specific objective may force us to employ techniques that can aggravate other objectives. 
For instance, one prominent technique to mitigate soft errors in a computing hardware is through redundancy/re-execution approach, i.e., performing a task multiple times and select an appropriate result in the end through voting. 
This technique can be employed to improve the reliability of a system, but it significantly incurs more processing time and power/energy. 
Toward this, in our perspective, it is important to have and employ a synergistic development framework for building neuromorphic-based robotic systems, that maximizes benefits in terms of performance, reliability, and security.

\vspace{0.1cm}
\textit{\underline{Research developments and opportunities:}} 
To summarize, we list some new opportunities that arise from the requirements of a development framework for neuromorphic-based robots. 
\begin{itemize}
    \item The framework should facilitate multi-objective functions to maximize trade-off benefits among performance (i.e., learning quality, adaptability, and efficiency), reliability, and security. 
    \item The framework should encompass both HW- and SW-layer developments, across the full processing pipeline of a robotic system (i.e., sensors, computation, and actuation), hence enabling a synergistic solution. 
    \item An integrated development tool is required for creating and deploying SNNs for the given application use-cases, thereby streamlining the prototyping of neuromorphic-based robots.
\end{itemize}

%%%%%%%%%%%%%%%%%%%%%%%%%%%%%%%%%%%%%%%%%%%%%%%%%%%%%%%%%%%%%%%%%%%%%%%%%%%%%%%%
%%%%%%%%%%%%%%%%%%%%%%%%%%%%%%%%%%%%%%%%%%%%%%%%%%%%%%%%%%%%%%%%%%%%%%%%%%%%%%%%
\section{Conclusion}
\label{Sec_Conclude}

In this paper, we elaborate new challenges and opportunities that arise from the emergence of neuromorphic AI-based robotics field. 
We also present our perspectives on different aspects that have important roles for moving the field forward. 
First, learning quality and adaptability are essential for realizing the neuromorphic intelligence. 
Second, a synergistic development encompassing optimizations, fault mitigation, and defense mechanism, is crucial to ensure efficient and robust SNN processing. 
Third, representative and fair benchmarks are needed for continuous developments of the field. 
The preliminary empirical results align with our perspectives, thereby providing practical guidelines for future research developments toward enabling neuromorphic AI-based robotics.

% \label{Sec_Appendix}

% ...

%%%%%%%%%%%%%%%%%%%%%%%%%%%%%%%%%%%%%%%%%%%%%%%%%%%%%%%%%%%%%%%%%%%%%%%%%%%%%%%%
%%%%%%%%%%%%%%%%%%%%%%%%%%%%%%%%%%%%%%%%%%%%%%%%%%%%%%%%%%%%%%%%%%%%%%%%%%%%%%%%
\section*{Acknowledgment}
\label{Sec_Ack}
 
This work was partially supported by the NYUAD Center for Interacting Urban Networks (CITIES), funded by Tamkeen under the NYUAD Research Institute Award CG001.

%%%%%%%%%%%%%%%%%%%%%%%%%%%%%%%%%%%%%%%%%%%%%%%%%%%%%%%%%%%%%%%%%%%%%%%%%%%%%%%%
%%%%%%%%%%%%%%%%%%%%%%%%%%%%%%%%%%%%%%%%%%%%%%%%%%%%%%%%%%%%%%%%%%%%%%%%%%%%%%%%
\bibliographystyle{IEEEtran}
\bibliography{bibliography.bib}

\begin{thebibliography}{10}
\providecommand{\url}[1]{#1}
\csname url@samestyle\endcsname
\providecommand{\newblock}{\relax}
\providecommand{\bibinfo}[2]{#2}
\providecommand{\BIBentrySTDinterwordspacing}{\spaceskip=0pt\relax}
\providecommand{\BIBentryALTinterwordstretchfactor}{4}
\providecommand{\BIBentryALTinterwordspacing}{\spaceskip=\fontdimen2\font plus
\BIBentryALTinterwordstretchfactor\fontdimen3\font minus
  \fontdimen4\font\relax}
\providecommand{\BIBforeignlanguage}[2]{{%
\expandafter\ifx\csname l@#1\endcsname\relax
\typeout{** WARNING: IEEEtran.bst: No hyphenation pattern has been}%
\typeout{** loaded for the language `#1'. Using the pattern for}%
\typeout{** the default language instead.}%
\else
\language=\csname l@#1\endcsname
\fi
#2}}
\providecommand{\BIBdecl}{\relax}
\BIBdecl

\bibitem{Ref_Bartolozzi_EmbodiedNeuroIntel_Nature22}
C.~Bartolozzi, G.~Indiveri, and E.~Donati, ``Embodied neuromorphic
  intelligence,'' \emph{Nature communications}, vol.~13, no.~1, p. 1024, 2022.

\bibitem{Ref_Roy_SpikeMachineIntel_Nature19}
K.~Roy, A.~Jaiswal, and P.~Panda, ``Towards spike-based machine intelligence
  with neuromorphic computing,'' \emph{Nature}, vol. 575, no. 7784, pp.
  607--617, 2019.

\bibitem{Ref_Schuman_OpportunityNeuro_Nature22}
C.~D. Schuman \emph{et~al.}, ``Opportunities for neuromorphic computing
  algorithms and applications,'' \emph{Nature Computational Science}, vol.~2,
  no.~1, pp. 10--19, 2022.

\bibitem{Ref_Putra_FSpiNN_TCAD20}
R.~V.~W. {Putra} and M.~{Shafique}, ``Fspinn: An optimization framework for
  memory-efficient and energy-efficient spiking neural networks,'' \emph{IEEE
  Trans. on Computer-Aided Design of Integrated Circuits and Systems (TCAD)},
  vol.~39, no.~11, pp. 3601--3613, 2020.

\bibitem{Ref_Putra_lpSpikeCon_IJCNN22}
R.~V.~W. Putra and M.~Shafique, ``lpspikecon: Enabling low-precision spiking
  neural network processing for efficient unsupervised continual learning on
  autonomous agents,'' in \emph{Int. Joint Conf. on Neural Networks (IJCNN)},
  2022, pp. 1--8.

\bibitem{Ref_Rathi_SNNsurvey_CSUR23}
N.~Rathi \emph{et~al.}, ``Exploring neuromorphic computing based on spiking
  neural networks: Algorithms to hardware,'' \emph{ACM Comput. Surv. (CSUR)},
  vol.~55, no.~12, March 2023.

\bibitem{Ref_Sironi_HATS_CVPR18}
A.~Sironi \emph{et~al.}, ``Hats: Histograms of averaged time surfaces for
  robust event-based object classification,'' in \emph{IEEE Conf. on Computer
  Vision and Pattern Recognition (CVPR)}, 2018, pp. 1731--1740.

\bibitem{Ref_Safa_CameraRadarSLAMsnn_ICRA23}
A.~Safa \emph{et~al.}, ``Fusing event-based camera and radar for slam using
  spiking neural networks with continual stdp learning,'' in \emph{2023 IEEE
  Int. Conf. on Robotics and Automation (ICRA)}, 2023, pp. 2782--2788.

\bibitem{Ref_Diehl_STDPmnist_FNCOM15}
P.~Diehl and M.~Cook, ``Unsupervised learning of digit recognition using
  spike-timing-dependent plasticity,'' \emph{Frontiers in Computational
  Neuroscience}, vol.~9, p.~99, 2015.

\bibitem{Ref_Marchisio_RSNN_IROS21}
A.~Marchisio \emph{et~al.}, ``R-snn: An analysis and design methodology for
  robustifying spiking neural networks against adversarial attacks through
  noise filters for dynamic vision sensors,'' in \emph{IEEE/RSJ Int. Conf. on
  Intelligent Robots and Systems (IROS)}, 2021, pp. 6315--6321.

\bibitem{Ref_Putra_SpikeDyn_DAC21}
R.~V.~W. Putra and M.~Shafique, ``Spikedyn: A framework for energy-efficient
  spiking neural networks with continual and unsupervised learning capabilities
  in dynamic environments,'' in \emph{58th ACM/IEEE Design Automation Conf.
  (DAC)}, 2021, pp. 1057--1062.

\bibitem{Ref_Basu_SNNicSurvey_CICC22}
A.~Basu \emph{et~al.}, ``Spiking neural network integrated circuits: A review
  of trends and future directions,'' in \emph{IEEE Custom Integrated Circuits
  Conf. (CICC)}, 2022, pp. 1--8.

\bibitem{Zhu10419072}
N.~Zhu, Z.~Xi, C.~Wu, F.~Zhong, R.~Qi, H.~Chen, S.~Xu, and W.~Ji, ``Inductive
  conformal prediction enhanced lstm-snn network: Applications to birds and
  uavs recognition,'' \emph{IEEE Geoscience and Remote Sensing Letters},
  vol.~21, pp. 1--5, 2024.

\bibitem{10.3389/fnins.2021.667011}
L.~Steffen \emph{et~al.}, ``Benchmarking highly parallel hardware for spiking
  neural networks in robotics,'' \emph{Frontiers in Neuroscience}, vol.~15,
  2021.

\bibitem{Kreiser8594228}
R.~Kreiser \emph{et~al.}, ``Pose estimation and map formation with spiking
  neural networks: towards neuromorphic slam,'' in \emph{IEEE/RSJ Int. Conf. on
  Intelligent Robots and Systems (IROS)}, 2018, pp. 2159--2166.

\bibitem{Kreiser9197498}
------, ``Error estimation and correction in a spiking neural network for map
  formation in neuromorphic hardware,'' in \emph{IEEE Int. Conf. on Robotics
  and Automation (ICRA)}, 2020, pp. 6134--6140.

\bibitem{zhang2023dynamic}
X.~Zhang \emph{et~al.}, ``Dynamic obstacle avoidance for unmanned aerial
  vehicle using dynamic vision sensor,'' in \emph{Int. Conf. on Artificial
  Neural Networks (ICANN)}.\hskip 1em plus 0.5em minus 0.4em\relax Springer,
  2023, pp. 161--173.

\bibitem{Yu10365576_2023}
W.~Yu \emph{et~al.}, ``Fault-tolerant attitude tracking control driven by
  spiking nns for unmanned aerial vehicles,'' \emph{IEEE Trans. on Neural
  Networks and Learning Systems (TNNLS)}, pp. 1--13, 2023.

\bibitem{Ref_Viale_CarSNN_IJCNN21}
A.~Viale \emph{et~al.}, ``Carsnn: An efficient spiking neural network for
  event-based autonomous cars on the loihi neuromorphic research processor,''
  in \emph{Int. Joint Conf. on Neural Networks (IJCNN)}, 2021, pp. 1--10.

\bibitem{BING202021}
Z.~Bing \emph{et~al.}, ``Indirect and direct training of spiking neural
  networks for end-to-end control of a lane-keeping vehicle,'' \emph{Neural
  Networks}, vol. 121, pp. 21--36, 2020.

\bibitem{10.1007/978-3-319-46687-3_21}
R.~de~Azambuja \emph{et~al.}, ``Graceful degradation under noise on brain
  inspired robot controllers,'' in \emph{Neural Information Processing}.\hskip
  1em plus 0.5em minus 0.4em\relax Cham: Springer International Publishing,
  2016, pp. 195--204.

\bibitem{doi:10.1126/sciadv.abl5068}
I.~Krauhausen \emph{et~al.}, ``Organic neuromorphic electronics for
  sensorimotor integration and learning in robotics,'' \emph{Science Advances},
  vol.~7, no.~50, p. eabl5068, 2021.

\bibitem{Ref_Rueckauer_ANNtoSNN_FNINS17}
B.~Rueckauer \emph{et~al.}, ``Conversion of continuous-valued deep networks to
  efficient event-driven networks for image classification,'' \emph{Frontiers
  in Neuroscience (FNINS)}, vol.~11, p. 682, 2017.

\bibitem{Ref_Neftci_SurrogateSNNs_IEEEMSP19}
E.~O. Neftci, H.~Mostafa, and F.~Zenke, ``Surrogate gradient learning in
  spiking neural networks: Bringing the power of gradient-based optimization to
  spiking neural networks,'' \emph{IEEE Signal Processing Magazine}, vol.~36,
  no.~6, pp. 51--63, 2019.

\bibitem{Ref_Kaiser_DECOLLE_FNINS20}
J.~Kaiser \emph{et~al.}, ``Synaptic plasticity dynamics for deep continuous
  local learning (decolle),'' \emph{Frontiers in Neuroscience}, vol.~14, 2020.

\bibitem{Ref_Putra_Mantis_ICARA23}
R.~V.~W. Putra and M.~Shafique, ``Mantis: Enabling energy-efficient autonomous
  mobile agents with spiking neural networks,'' in \emph{9th Int. Conf. on
  Automation, Robotics and Applications}, 2023, pp. 197--201.

\bibitem{Ref_Putra_QSpiNN_IJCNN21}
------, ``Q-spinn: A framework for quantizing spiking neural networks,'' in
  \emph{Int. Joint Conf. on Neural Networks (IJCNN)}, 2021, pp. 1--8.

\bibitem{Ref_Sen_ApproxSNN_DATE17}
S.~{Sen} \emph{et~al.}, ``Approximate computing for spiking neural networks,''
  in \emph{DATE}, March 2017, pp. 193--198.

\bibitem{Ref_Chowdhury_LowLatencySNNs_ECCV22}
S.~S. Chowdhury, N.~Rathi, and K.~Roy, ``Towards ultra low latency spiking
  neural networks for vision and sequential tasks using temporal pruning,'' in
  \emph{European Conf. on Computer Vision (ECCV)}.\hskip 1em plus 0.5em minus
  0.4em\relax Springer, 2022, pp. 709--726.

\bibitem{Ref_Putra_TopSpark_IROS23}
R.~V.~W. Putra and M.~Shafique, ``Topspark: A timestep optimization methodology
  for energy-efficient spiking neural networks on autonomous mobile agents,''
  in \emph{IEEE/RSJ Int. Conf. on Intelligent Robots and Systems (IROS)}, 2023,
  pp. 3561--3567.

\bibitem{Ref_Na_AutoSNN_ICML22}
B.~Na \emph{et~al.}, ``Autosnn: Towards energy-efficient spiking neural
  networks,'' in \emph{Int. Conf. on Machine Learning}, 2022, pp.
  16\,253--16\,269.

\bibitem{Ref_Kim_SNASNet_ECCV22}
Y.~Kim \emph{et~al.}, ``Neural architecture search for spiking neural
  networks,'' in \emph{European Conf. on Computer Vision (ECCV)}, 2022, pp.
  36--56.

\bibitem{Ref_Putra_SpikeNAS_arXiv24}
R.~V.~W. Putra and M.~Shafique, ``Spikenas: A fast memory-aware neural
  architecture search framework for spiking neural network systems,''
  \emph{arXiv preprint arXiv:2402.11322}, 2024.

\bibitem{Ref_Achararit_APNAS_Access20}
P.~Achararit \emph{et~al.}, ``Apnas: Accuracy-and-performance-aware neural
  architecture search for neural hardware accelerators,'' \emph{IEEE Access},
  vol.~8, pp. 165\,319--165\,334, 2020.

\bibitem{Ref_Krithivasan_SpikeBundle_ISLPED19}
S.~{Krithivasan} \emph{et~al.}, ``Dynamic spike bundling for energy-efficient
  spiking neural networks,'' in \emph{IEEE/ACM Int. Symp. on Low Power
  Electronics and Design (ISLPED)}, July 2019, pp. 1--6.

\bibitem{Ref_Putra_SparkXD_DAC21}
R.~V.~W. Putra, M.~A. Hanif, and M.~Shafique, ``Sparkxd: A framework for
  resilient and energy-efficient spiking neural network inference using
  approximate dram,'' in \emph{58th ACM/IEEE Design Automation Conference
  (DAC)}, 2021, pp. 379--384.

\bibitem{Ref_Srinivasan_SLSTDP_IJCNN17}
G.~Srinivasan \emph{et~al.}, ``Spike timing dependent plasticity based enhanced
  self-learning for efficient pattern recognition in spiking neural networks,''
  in \emph{Int. Joint Conf. on Neural Networks}, 2017, pp. 1847--1854.

\bibitem{Ref_Putra_EnforceSNN_FNINS22}
R.~V.~W. Putra, M.~A. Hanif, and M.~Shafique, ``Enforcesnn: Enabling resilient
  and energy-efficient spiking neural network inference considering approximate
  drams for embedded systems,'' \emph{Frontiers in Neuroscience (FNINS)},
  vol.~16, p. 937782, 2022.

\bibitem{Ref_Putra_DRMap_DAC20}
------, ``Drmap: A generic dram data mapping policy for energy-efficient
  processing of convolutional neural networks,'' in \emph{57th ACM/IEEE Design
  Automation Conf. (DAC)}, 2020, pp. 1--6.

\bibitem{Ref_Putra_ROMANet_TVLSI21}
------, ``Romanet: Fine-grained reuse-driven off-chip memory access management
  and data organization for deep neural network accelerators,'' \emph{IEEE
  Trans. on Very Large Scale Integration Systems (TVLSI)}, vol.~29, no.~4, pp.
  702--715, 2021.

\bibitem{Ref_Asifuzzaman_SurveyPIM_Memori23}
K.~Asifuzzaman \emph{et~al.}, ``A survey on processing-in-memory techniques:
  Advances and challenges,'' \emph{Memories-Materials, Devices, Circuits and
  Systems (Memori)}, vol.~4, p. 100022, 2023.

\bibitem{Ref_Shafique_EdgeAI_ICCAD21}
M.~Shafique \emph{et~al.}, ``Towards energy-efficient and secure edge ai: A
  cross-layer framework iccad special session paper,'' in \emph{IEEE/ACM Int.
  Conf. On Computer Aided Design (ICCAD)}, 2021, pp. 1--9.

\bibitem{Ref_Putra_SoftSNN_DAC22}
R.~V.~W. Putra, M.~A. Hanif, and M.~Shafique, ``Softsnn: Low-cost fault
  tolerance for spiking neural network accelerators under soft errors,'' in
  \emph{59th ACM/IEEE Design Automation Conf. (DAC)}, 2022, pp. 151--156.

\bibitem{brockman2016openai}
G.~Brockman \emph{et~al.}, ``Openai gym,'' \emph{arXiv preprint:1606.01540},
  2016.

\bibitem{krotkov2018darpa}
E.~Krotkov \emph{et~al.}, ``The darpa robotics challenge finals: Results and
  perspectives,'' \emph{The DARPA robotics challenge finals: Humanoid robots to
  the rescue}, pp. 1--26, 2018.

\bibitem{Andrew2013low}
Andrew \emph{et~al.}, ``Low cost localisation for agricultural robotics,'' in
  \emph{Australasian Conf. on Robotics and Automation}, 2013, pp. 1--8.

\bibitem{kumar2024robohive}
V.~Kumar \emph{et~al.}, ``Robohive: A unified framework for robot learning,''
  \emph{Advances in Neural Information Processing Systems (NeurIPS)}, vol.~36,
  2024.

\bibitem{Gallego_2022TPAMI_EventBasedVisionSurvey}
G.~Gallego \emph{et~al.}, ``Event-based vision: {A} survey,'' \emph{{IEEE}
  Trans. Pattern Anal. Mach. Intell. (TPAMI)}, vol.~44, no.~1, pp. 154--180,
  2022.

\bibitem{schoepe2023event}
T.~Schoepe \emph{et~al.}, ``Event-based sound source localization in
  neuromorphic systems,'' \emph{Authorea Preprints}, 2023.

\bibitem{muthusamy2020neuromorphic}
R.~Muthusamy \emph{et~al.}, ``Neuromorphic event-based slip detection and
  suppression in robotic grasping and manipulation,'' \emph{IEEE Access},
  vol.~8, pp. 153\,364--153\,384, 2020.

\bibitem{cadena2016past}
C.~Cadena \emph{et~al.}, ``Past, present, and future of simultaneous
  localization and mapping: Toward the robust-perception age,'' \emph{IEEE
  Trans. on Robotics (TRO)}, vol.~32, no.~6, pp. 1309--1332, 2016.

\bibitem{xing2019firenose}
Y.~Xing \emph{et~al.}, ``Firenose on mobile robot in harsh environments,''
  \emph{IEEE Sensors Journal (JSEN)}, vol.~19, no.~24, pp. 12\,418--12\,431,
  2019.

\bibitem{hasib2021comprehensive}
S.~A. Hasib \emph{et~al.}, ``A comprehensive review of available battery
  datasets, rul prediction approaches, and advanced battery management,''
  \emph{IEEE Access}, vol.~9, pp. 86\,166--86\,193, 2021.

\bibitem{Ref_Putra_ReSpawn_ICCAD21}
R.~V.~W. Putra, M.~A. Hanif, and M.~Shafique, ``Respawn: Energy-efficient
  fault-tolerance for spiking neural networks considering unreliable
  memories,'' in \emph{IEEE/ACM Int. Conf. On Computer Aided Design (ICCAD)},
  2021, pp. 1--9.

\bibitem{Ref_Putra_RescueSNN_FNINS23}
------, ``Rescuesnn: enabling reliable executions on spiking neural network
  accelerators under permanent faults,'' \emph{Frontiers in Neuroscience
  (FNINS)}, vol.~17, 2023.

\bibitem{Ref_ElAllami_SecuringSNNs_DATE21}
R.~El-Allami \emph{et~al.}, ``Securing deep spiking neural networks against
  adversarial attacks through inherent structural parameters,'' in
  \emph{Design, Automation \& Test in Europe Conf. \& Exhibition}, 2021, pp.
  774--779.

\bibitem{Marchisio_2021IROS_RSNN}
A.~Marchisio \emph{et~al.}, ``R-snn: An analysis and design methodology for
  robustifying spiking neural networks against adversarial attacks through
  noise filters for dynamic vision sensors,'' in \emph{IEEE/RSJ Int. Conf. on
  Intelligent Robots and Systems (IROS)}, 2021, pp. 6315--6321.

\bibitem{Kim_2022AAAI_PrivateSNN}
Y.~Kim, Y.~Venkatesha, and P.~Panda, ``Privatesnn: Privacy-preserving spiking
  neural networks,'' in \emph{Thirty-Sixth {AAAI} Conf. on Artificial
  Intelligence (AAAI)}.\hskip 1em plus 0.5em minus 0.4em\relax {AAAI} Press,
  2022, pp. 1192--1200.

\bibitem{Nikfam_2023Information_HESNN}
F.~Nikfam \emph{et~al.}, ``A homomorphic encryption framework for
  privacy-preserving spiking neural networks,'' \emph{Inf.}, vol.~14, no.~10,
  p. 537, 2023.

\bibitem{Ref_Marchisio_IsSpikingSecure_IJCNN20}
A.~Marchisio \emph{et~al.}, ``Is spiking secure? a comparative study on the
  security vulnerabilities of spiking and deep neural networks,'' in \emph{Int.
  Joint Conf. on Neural Networks (IJCNN)}, 2020, pp. 1--8.

\bibitem{Ref_Venceslai_NeuroAttack_IJCNN20}
V.~Venceslai \emph{et~al.}, ``Neuroattack: Undermining spiking neural networks
  security through externally triggered bit-flips,'' in \emph{Int. Joint Conf.
  on Neural Networks (IJCNN)}, 2020, pp. 1--8.

\bibitem{Marchisio_2021IJCNN_DVSAttacks}
A.~Marchisio \emph{et~al.}, ``Dvs-attacks: Adversarial attacks on dynamic
  vision sensors for spiking neural networks,'' in \emph{Int. Joint Conf. on
  Neural Networks (IJCNN)}, 2021, pp. 1--9.

\bibitem{Dave_2022VTS_AgileMethodology}
S.~Dave \emph{et~al.}, ``Special session: Towards an agile design methodology
  for efficient, reliable, and secure {ML} systems,'' in \emph{40th {IEEE}
  {VLSI} Test Symposium (VTS)}, 2022, pp. 1--14.

\bibitem{Marchisio_2022Access_RoHNAS}
A.~Marchisio \emph{et~al.}, ``Rohnas: {A} neural architecture search framework
  with conjoint optimization for adversarial robustness and hardware efficiency
  of convolutional and capsule networks,'' \emph{{IEEE} Access}, vol.~10, pp.
  109\,043--109\,055, 2022.

\end{thebibliography}
\end{spacing}

\end{document}